\documentclass[letterpaper, 10 pt, conference]{ieeeconf}   % Comment this line out if you need a4paper

\IEEEoverridecommandlockouts                              % This command is only needed if 
\usepackage{graphics} % for pdf, bitmapped graphics files
\usepackage{epsfig} % for postscript graphics files
\usepackage{amsmath} % assumes amsmath package installed
\usepackage{svg}
\usepackage{stfloats}
\usepackage{url}
\usepackage{verbatim}
\usepackage{graphicx}
\usepackage{cite}
\usepackage{amssymb}
\usepackage{booktabs}
\usepackage{makecell}
\usepackage{threeparttable}

\usepackage{mathrsfs}

\title{\LARGE \bf
DenserRadar: A 4D millimeter-wave radar point cloud detector based on dense LiDAR point clouds
}

\author{Zeyu Han, Junkai Jiang, Xiaokang Ding, Qingwen Meng, Shaobing Xu, Lei He$^{*}$, Jianqiang Wang$^{*}$
\thanks{This work was supported by the National Natural Science Foundation of China(No. 52372415), the National Natural Science Foundation of China (Grant No. 52131201) and the Tsinghua-Toyota Joint Research Institute Inter-disciplinary Program.}% <-this % stops a space
\thanks{All the authors are with School of Vehicle and Mobility, Tsinghua University, Beijing, China}
\thanks{$^{*}$Correspondence:   {\tt helei2023@tsinghua.edu.cn} (L.H.), {\tt wjqlws@tsinghua.edu.cn} (J.W.)}
}

\begin{document}

\maketitle
\thispagestyle{empty}
\pagestyle{empty}

\begin{abstract}
% 4D Radar的作用+点云稀疏性的问题+本文提出的方法+效果+解决的问题
The 4D millimeter-wave (mmWave) radar, with its robustness in extreme environments, extensive detection range, and capabilities for measuring velocity and elevation, has demonstrated significant potential for enhancing the perception abilities of autonomous driving systems in corner-case scenarios. Nevertheless, the inherent sparsity and noise of 4D mmWave radar point clouds restrict its further development and practical application. In this paper, we introduce a novel 4D mmWave radar point cloud detector, which leverages high-resolution dense LiDAR point clouds. Our approach constructs dense 3D occupancy ground truth from stitched LiDAR point clouds, and employs a specially designed network named DenserRadar. The proposed method surpasses existing probability-based and learning-based radar point cloud detectors in terms of both point cloud density and accuracy on the K-Radar dataset.

\end{abstract}

% \begin{IEEEkeywords}
% 4D millimeter-wave radar,  Autonomous driving, Point cloud detection.
% \end{IEEEkeywords}

\section{Introduction} \label{Sec:Introduction}

% 自动驾驶很重要
Autonomous driving technology, which aims to provide safe, convenient, and comfortable transportation experiences, is advancing at an impressive pace. To realize high-level autonomous driving, the capabilities of sophisticated environment perception and localization are indispensable. Consequently, the sensors equipped on autonomous vehicles, including cameras, LiDARs, and radars, as well as their associated algorithms, are attracting increasing research interest. 

% 毫米波雷达很重要，介绍新兴的4D毫米波雷达
Acknowledging the advantages of compact size, cost efficiency, all-weather adaptation, velocity-measuring capability, and extensive detection range, etc.\cite{10078429}, mmWave radars have been widely employed in autonomous driving. The recent advancement in multiple-input multiple-output (MIMO) antenna technology has further improved the elevation resolution, leading to the emergence of the 4D mmWave radar. Consequently, the 4D mmWave radar is increasingly viewed as a pivotal enhancement for the perception and localization capabilities in autonomous driving, particularly in challenging corner-case scenarios such as rainy, snowy, foggy weathers. As its name suggests, the 4D mmWave radar is capable of measuring four dimensions of target information: range, azimuth, elevation, and Doppler velocity, offering a comprehensive sensing solution. 

% 4D毫米波雷达现存的问题，尤其是分辨率方面
However, the quality of 4D mmWave radar point clouds significantly falls behind that of LiDAR point clouds. First, 4D mmWave radar point clouds suffer from low resolution, especially in angular measurement. This limitation is mainly due to the radar's antenna configuration and Direction of Arrival (DOA) estimation\cite{8828025}. Second, 4D mmWave radar point clouds are much sparser than LiDAR point clouds. Third, 4D mmWave radar point clouds often contain numerous clutter points on account of multi-path effect, signal interference and ground reflection. All these shortcomings prevent the utilization of the 4D mmWave radar in autonomous driving.

% 分析radar点云生成的原理，可以发现除了硬件限制外，在信号处理流程中的CFAR对其点云质量影响也比较大。（简单说一下CFAR是干什么的）因此，本研究打算从优化点云提取算法入手，提高radar点云质量。
The quality of 4D mmWave radar point clouds is restricted by not only the hardware, bud also the signal processing algorithms\cite{han20234d}. In particular, detecting actual targets to generate point clouds from raw radar maps or tensors may greatly influence the quality. Traditionally, the Constant False Alarm Rate(CFAR) detector and its variants\cite{shor1991performances, raghavan1992analysis} are widely applied to detect radar point clouds. However, as probability-based algorithms, CFAR-type detectors may encounter problems in detecting objects with varying sizes, as they are not independent and identically distributed\cite{cheng2022novel}, which is frequently appears in autonomous driving contexts.

% 为解决现存问题，本研究的方法流程和主要贡献
To address the point cloud quality issues associated with the 4D mmWave radar, this paper proposes a learning-based 4D mmWave radar point cloud detector supervised by dense ground truth information generated from LiDAR point clouds. Initially, we stitch multiple frames of pre-processed LiDAR point clouds to generate dense 3D occupancy ground truth. Then we introduce the DenserRadar network, which extracts feature of raw 4D mmWave radar tensors, generating 4D mmWave radar point clouds with higher density and accuracy. A weighted hybrid loss function, among other novel design elements, to capture multi-resolution features and generate point clouds with superior resolution compared to conventional techniques. Comparative experiments on the K-Radar dataset\cite{paek2022k} demonstrate the efficacy of our approach. The contributions of this work are outlined as follows:

\begin{itemize}
    \item{Our work is the first 4D mmWave radar point cloud detector that is supervised by dense 3D occupancy ground truth, which is generated from stitched multiple frames LiDAR point clouds, thereby densifying the detected radar point clouds.}
    \item{We propose an innovative dense 3D occupancy ground truth generation pipeline, and the stitched dense LiDAR point clouds of the K-Radar dataset, which provide comprehensive scene ground truth, will be made accessible upon publication to support further research.}
    \item{Owing to the specialized designs of the DenserRadar network, our algorithm outperforms existing CFAR-type and learning-based radar point cloud detection methods in terms of both density and accuracy.}
\end{itemize}

\section{Related Works}  \label{Sec:Related}
As an emerging sensor in the field of autonomous driving, there has been a number of 4D mmWave radar-related studies, including its perception and localization algorithms, as well as point cloud detection methods. This section will give a brief revisiting of the relevant literature.
% 简单说一说感知定位上的现有研究
\subsection{4D mmWave Radar in Autonomous Driving}
Benefiting from its perception ability under low visibility and adverse weather conditions, the 4D mmWave radar is recognized as a robust sensor on object detection\cite{tan20223d, yan2023mvfan, paek2023enhanced} and tracking\cite{pan2023moving, liu2023framework}. Most researchers adapt LiDAR-based algorithms for 4D mmWave radar point cloud while leveraging its advantages. Tan et al. \cite{tan20223d} compute ego-velocity by Doppler information to stitch historical frames and enhance object detection. The Doppler information is also integrated into position embeddings by Yan et al.\cite{yan2023mvfan} for more accurate feature extraction. Instead of radar point cloud, the pre-CFAR 4D tensor is utilized by Paek et al.\cite{paek2023enhanced} for Bird-Eye-View(BEV) 2D object detection.  

As for localization, some 4D mmWave radar Simultaneous Localization and Mapping(SLAM) systems have been proposed\cite{zhuang20234d, zhang20234dradarslam, li20234d}. Researchers commonly employs Doppler information for ego-velocity estimation and dynamic points removal on the basis of traditional LiDAR SLAM algorithms.

% 传统：各种CFAR；学习：……，说他们的不足反映本文创新
\subsection{4D mmWave Radar Point Cloud Generation}
Traditional mmWave radar point cloud generation methods mostly apply the probability-guided CFAR algorithm, which sets an intensity threshold based on background noise to filter out detected points. Variations of CFAR exist, contingent on the method of background noise estimation. The most widely adopted Cell Averaging(CA)-CFAR\cite{raghavan1992analysis} calculates the background noise $Z_{CA}$, as the average value of the cells outside of the guard cells on both sides of the detected cell. A cell is identified as a real object if its intensity value $Z_d$ satisfies $Z_d > \delta_{CA} Z_{CA}$, where $\delta_{CA}$ represents  the threshold of CA-CFAR. Similarly, the Ordered Statistics(OS)-CFAR\cite{shor1991performances} determines the background noise $Z_{OS}$ by selecting the $k$-th smallest value of the cells outside of the guard cells on both sides of the detected cell, and a cell is classified as a real object if $Z_d > \delta_{OS} Z_{OS}$, with $\delta_{OS}$ being the threshold of OS-CFAR.
% \begin{equation}
%     Z_{CA} = \frac{1}{2N} (\sum_{i=1}^{N}Z^{l}_{i}+\sum_{i=1}^{N}Z^{r}_{i}),
% \end{equation}
% And Ordered Statistics(OS)-CFAR\cite{shor1991performances} selects the $k$-th smallest value of the cells on both sides as the background noise:
% \begin{align}
%     Z^{*}_{1} \leq Z^{*}_{2} \leq ... \leq Z^{*}_{k}  \leq ...& \leq Z^{*}_{2N-1} \leq Z^{*}_{2N}, \\
%     Z_{OS} = & Z^{*}_{k}.
% \end{align}

It is evident that CFAR-type algorithms primarily consider the intensity values of a select number of cells, neglecting the Doppler information, let alone the whole scene information, causing inevitable information lost. To handle this problem, several researchers propose learning-based mmWave radar point cloud detectors. Brodeski et al. \cite{brodeski2019deep} firstly apply a Convolutional Neural Networks(CNN)-based segmentation network on Range-Doppler(RD) maps, using ground truth data obtained through a calibration process in an anechoic chamber. Although the scenarios are limited, their work proves the validity of learning-based radar point cloud detection. Furthermore, Some researchers have explored the use of LiDAR point clouds for supervision, developing various network architectures such as U-Net-based \cite{cheng2021new}, Generative Adversarial Networks(GAN)-based\cite{cheng2022novel} and CNN-based\cite{lu2023novel} architectures models to refine the detection of radar point clouds.

Upon examining existing learning-based mmWave radar point cloud detectors, it becomes apparent that the main drawback of these algorithms is that the inadequate supervisory data (such as single-frame LiDAR point clouds) leads to insufficient density of the detected radar point clouds. Additionally, the designed networks are also incapable for improving the resolution of radar point clouds. To overcome these challenges, this study proposes a novel approach that stitches multiple frames of LiDAR point clouds to provide enriched supervision, and constructs the DenserRadar network to improve the density of detected radar point clouds.

\section{Methodology}  \label{Sec:Methodology}
This section will firstly provide a brief overview of our algorithm, and then introduce the details of ground truth generation pipeline and the architecture of the DenserRadar network.

\subsection{Overview}
\begin{figure}
    \centering
    \includegraphics[width = \linewidth]{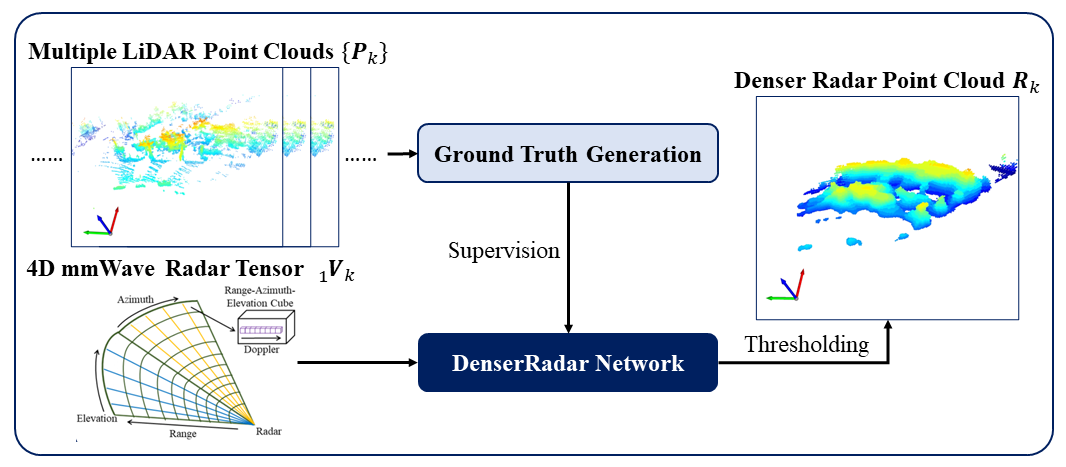}
    \caption{The overview of the whole algorithm.}
    \label{Fig:Overview}
\end{figure}
Our algorithm is illustrated in Fig. \ref{Fig:Overview}. We firstly design a ground truth generation pipeline to obtain dense 3D occupancy ground truth by stitching multiple frames of LiDAR point clouds as the supervision, and then establish the DenserRadar network, which is tasked with detecting radar point clouds from the raw 4D mmWave radar tensor data.

\subsection{Ground Truth Generation}

\begin{figure*}
    \centering
    \includegraphics[width = \linewidth]{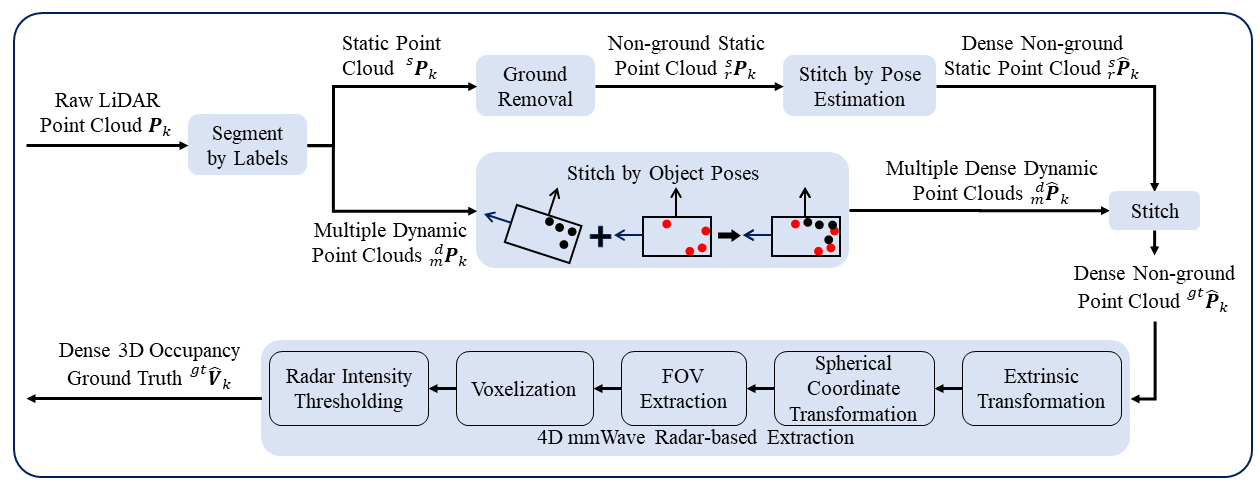}
    \caption{The ground truth generation pipeline.}
    \label{Fig:Gtgen}
\end{figure*}
The ground truth generation process is shown in Fig. \ref{Fig:Gtgen}. By leveraging the object labeling in the dataset, the $k$th frame of the raw LiDAR point cloud, denoted as $\textbf{P}_{k}=\{p_{i}\}_{k}, i=1,2,3,...n$, is partitioned into a static ($s$) point cloud $^{s}\textbf{P}_{k}$ and multiple dynamic ($d$) point clouds $^{d}_{m}\textbf{P}_{k}$ corresponding to individual objects, where $m$ denotes the tracking id of each object. 

For each frame of the static point cloud, we apply a random sample consensus(RANSAC)-based method \cite{fischler1981random} to fit the ground plane and remove ground points, resulting in a non-ground static point cloud $^{s}_{r}\textbf{P}_{k}$. Then the planar features and corner features are extracted by methods adapted from \cite{zhang2014loam} and these features are registered by a iterative closest point(ICP)-based algorithm\cite{bai2022icp} to estimate the relative pose drift $^{s}\textbf{T}_{k}^{k+1}$ 
between consecutive frames. Moreover, each $^{s}_{r}\textbf{P}_{k}$ is stitched with the static point clouds of the preceding and subsequent $t$ frames based on their respective poses, yielding a densified static point cloud $^{s}_{r}\hat{\textbf{P}}_{k}$, as illustrated by the following equation:

\begin{equation}
\begin{split}
    ^{s}_{r}\hat{\textbf{P}}_{k} = & \sum_{i=1}^{t}(\prod^{i}_{j=1} {^{s}\textbf{T}_{k-j}^{k-j+1}}) ^{s}_{r}\textbf{P}_{k-i} \\
    & + ^{s}_{r}\textbf{P}_{k} \\
    & + \sum_{i=1}^{t}(\prod^{i}_{j=1} {^{s}\textbf{T}_{k+j}^{k+j-1}}) ^{s}_{r}\textbf{P}_{k+i}.
\end{split}
\end{equation}

On the other hand, the dynamic point cloud $^{d}_{m}\textbf{P}_{k}$ corresponding to each object is also stitched with the preceding and subsequent $t$ frames of corresponding point clouds. Specifically, given the pose transformation matrix $^{d}_{m}\textbf{T}_{k}$, which represents the relative position of each object $m$ to the sensor at each frame, the stitched point cloud of object $m$ can be obtained as following:

\begin{equation}
\begin{split}
    ^{d}_{m}\hat{\textbf{P}}_{k} = & \sum_{i=1}^{t}({^{d}_{m}\textbf{T}_{k-i}} ({^{d}_{m}\textbf{P}_{k-i}}) {^{d}_{m}\textbf{T}_{k}}) \\
    & + ^{d}_{m}\textbf{P}_{k} \\
    & + \sum_{i=1}^{t}( {^{d}_{m}\textbf{T}_{k+i}^{T}}(^{d}_{m}\textbf{P}_{k+i}) {^{d}_{m}\textbf{T}_{k}}).
\end{split}
\end{equation}

% we align and concatenate the direction and position of the corresponding dynamic object's detection boxes from the preceding and following ten frames with those of the current frame using true value labels, resulting in a dense dynamic point cloud. 
The dense, ground-removed LiDAR point cloud of each frame, denoted as $^{gt}\hat{\textbf{P}}_{k}$, is attained by stitching the densified static point cloud and densified dynamic point cloud of each object, as follows:

\begin{equation}
    ^{gt}\hat{\textbf{P}}_{k} = ^{s}_{r}\hat{\textbf{P}}_{k} + \sum_{m=1}^{M}{^{d}_{m}\hat{\textbf{P}}_{k}}
\end{equation}

Each $^{gt}\hat{\textbf{P}}_{k}$ is transformed into the 4D mmWave radar coordinate system according to the extrinsic calibration parameters, subsequently converted into the spherical coordinate, and then truncated according to the detection range and Field-of-View(FOV) of the 4D mmWave radar. The point cloud is thereafter voxelized into 3D occupancy grid to match the data format of the 4D mmWave radar tensor. Notably, the occupancy grid is configured with double the range and angular resolution of the raw 4D mmWave radar tensor, which serves to enhance the density and resolution of the generated radar point cloud. Finally, the voxels are filtered based on the intensities of corresponding voxels in the 4D mmWave radar tensor, only those exceeding a predetermined threshold are preserved, yielding the final dense 3D occupancy ground truth $^{gt}\hat{\textbf{V}}_{k}$.

% 核心算法
\subsection{The DenserRadar Network Architecture}
\begin{figure*}
    \centering
    \includegraphics[width = \linewidth]{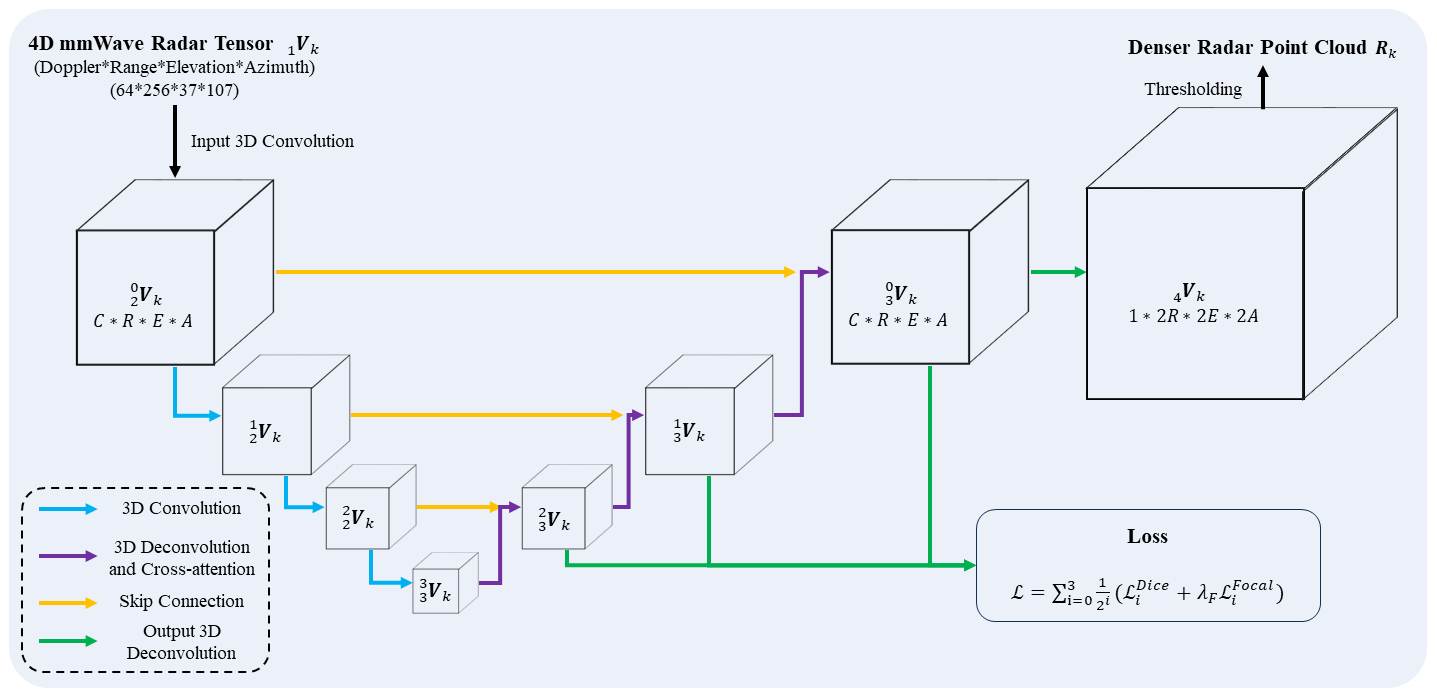}
    \caption{The DenserRadar network architecture. The size of each cube approximately represents its relative resolution.}
    \label{Fig:network}
\end{figure*}
Fig. \ref{Fig:network} represents the architecture of the proposed DenserRadar network. The network leverages a 3D U-net as the backbone, incorporating with input 3D convolution, output 3D deconvolution, and additional features such as a cross-attention mechanism and a novel design of loss function. The input 4D mmWave radar tensor of each frame $k$ is denoted as $_{1}\textbf{V}^{D \times R \times E \times A}_{k}$, where $D, R, E,$ and $A$ denoting the tensor's dimensions in Doppler, range, elevation and azimuth dimensions, respectively. 

\subsubsection{Input 3D Convolution}
The Doppler$(D)$ dimension in 4D mmWave radar tensors encodes the relative radial velocity information for each position in spherical coordinates, which has been shown to enhance angular resolutions\cite{chen2023drio}. Therefore, the Doppler dimension is interpreted as the channel dimension in the input tensor $_{1}\textbf{V}^{D \times R \times E \times A}_{k}$. After the initial 3D convolution, we obtain $^{0}_{2}\textbf{V}^{C \times R \times E \times A}_{k}$, where $C$ is the shape of the channel dimension. To preserve the dimensions of the range$(R)$, elevation $(E)$, and azimuth $(A)$, the kernel size for the convolution is set to $1 \times 1 \times 1$ in $R, E, A$ dimensions, respectively:

\begin{equation}
    ^{0}_{2}\textbf{V}^{C \times R \times E \times A}_{k} = 3DConv(_{1}\textbf{V}^{D \times R \times E \times A}_{k}).
\end{equation}

%TODO:扩充具体公式？
\subsubsection{3D U-net-based Backbone}
As a network initially designed for biomedical image segmentation, the U-net architecture\cite{ronneberger2015u} has demonstrated great capabilities in feature extraction and segmentation. To effectively extract features of pre-processed 4D mmWave radar tensor $^{0}_{2}\textbf{V}^{C \times R \times E \times A}_{k}$ in spherical coordinates from range, elevation and azimuth dimensions, this research proposes a 3D U-net architecture that includes three layers of 3D convolution and 3D deconvolution. Each of the convolution and deconvolution layer is designed to halve and double the dimensions of the input tensor, respectively. Considering the range resolution of the 4D mmWave radar is typically superior to that of the elevation and azimuth resolution, the sizes of convolution and deconvolution kernels are set to $5 \times 3 \times 3$ for the $R, E, A$ dimensions, respectively. The output of the backbone is denoted as $^{0}_{3}\textbf{V}^{C \times R \times E \times A}_{k}$:

\begin{equation}
    ^{0}_{3}\textbf{V}^{C \times R \times E \times A}_{k} = Backbone(^{0}_{2}\textbf{V}^{C \times R \times E \times A}_{k}).
\end{equation}

\subsubsection{Output 3D Deconvolution}
Given the comparatively lower resolution and point density of current 4D mmWave radar relative to high-definition LiDAR, the output module of DenserRadar network includes a 3D deconvolution layer that upsamples the output of each 3D U-net deconvolution layer, doubling its resolution in $R, E, A$ dimensions. To compute occupancy probabilities at each spherical coordinate position, represented as $_{4}\textbf{V}^{1 \times 2R \times 2E \times 2A}_{k}$, the channel dimension is compressed and a sigmoid layer is appended subsequent to the final 3D deconvolution operation:

\begin{equation}
    _{4}\textbf{V}^{1 \times 2R \times 2E \times 2A}_{k} = Sigmoid(3DDeconv(^{0}_{3}\textbf{V}^{C \times R \times E \times A}_{k})).
\end{equation}

% \subsubsection{Dilation Kernels}
% Dilated convolution is a specialized convolution operation that expands the receptive field of the convolution kernel by introducing a dilation rate, without increasing the parameter number and computational burden. To enhance the segmentation and point cloud detection performance, we integrate dilation kernels in 3D convolution and deconvolution layers of the 3D U-net-based backbone, setting the dilation rate $d=2$.

\subsubsection{Cross-attention Mechanism}
To enhance feature integration within the backbone network, attention weights are computed for each pair of feature maps $^{i}_{2}\textbf{V}_{k}$ and $^{i+1}_{3}\textbf{V}_{k} (i=0,1,2)$ by a cross-attention layer positioned prior to each deconvolution layer:

\begin{equation}
    ^{i+1}_{3}\textbf{V}_{k} = CrossAtten(_{3}^{i+1}\textbf{V}_{k}, ^{i}_{2}\textbf{V}_{k}), i=0,1,2.
\end{equation}

%TODO:扩充具体loss公式？
\subsubsection{Loss Function}
In typical U-net-like networks with multiple deep layers,  the supervision of lower-resolution layers may be less effective due to the extensive back-propagation path. One strategy to mitigate this issue is calculating losses at each deconvolution layer with different weights\cite{wei2023surroundocc}. Furthermore, given the disparity between the number of occupied and unoccupied positions in the occupancy supervision, we  employ a combination of dice loss \cite{huang2018robust} and focal loss\cite{lin2017focal} to ease this imbalance. Consequently, we introduce a weighted hybrid loss function, which is formulated as follows:
% weighted hybrid
\begin{equation}
    \mathcal{L} = \sum_{i=0}^{2}\frac{1}{2^i}(\mathcal{L}_{i}^{Dice}+\lambda_{F}\mathcal{L}_{i}^{Focal}),
\end{equation}

where $\lambda_F$ represents the balancing weight between the dice loss and focal loss.

\section{Experiments}  \label{Sec:Experiments}
In this section, we deploy our ground truth generation and radar point cloud detection algorithms to an existing dataset. We establish evaluation metrics to benchmark our algorithm against existing CFAR-type and learning-based methods, thereby demonstrating its detection efficacy. Subsequently, we conduct ablation studies to prove the efficacy of the specialized designs incorporated within our method.

% 数据集、参数、预处理等
\subsection{Experimental Setup}
Taking into account the data format, the K-Radar dataset is selected for our experiments.This dataset comprises 4D mmWave radar tensors of the RETINA-4ST radar, along with calibrated point clouds from a 64-beam Ouster LiDAR. The information of these two sensors are detailed in Table. \ref{Tab:setup}.

\begin{table}[!htbp]
    \centering
    \caption{Detailed sensor information in the K-Radar dataset}
    \resizebox{\linewidth}{!}{
    \begin{tabular}{ccc}
        \toprule
        Information &4D Radar &64-beam LiDAR\\
        \midrule
        Output Data &\makecell[c]{64$\times$256$\times$107$\times$37 \\ size 4D tensor} &\makecell[c]{131072 \\ 3D points} \\ 
        Doppler Resolution &0.06m/s &N/A \\
        Range Resolution &0.46m &0.1cm \\
        Elevation Resolution &1$^{\circ}$ &0.35$^{\circ}$ \\
        Azimuth Resolution &1$^{\circ}$ &0.18$^{\circ}$ \\
        Frame Per Second(FPS) &10 &10 \\
        \bottomrule
    \end{tabular}}
    \label{Tab:setup}
\end{table}

Regarding the implementation details of our algorithm, we set the number of preceding and subsequent LiDAR point cloud frames $t$ to 10 for fully utilizing of dense LiDAR point clouds. The balancing weight $\lambda_F$ between dice loss and focal loss is set to 700. The first 5 sequences in K-Radar, totally 3144 frames of 4D mmWave radar tensors and LiDAR point clouds are utilized, 80\% and 20\% of which are split for training and testing, respectively.

% 或许设计一些点云方面的评价指标？
\subsection{Evaluation Metrics}
To evaluate the generated radar point clouds against LiDAR supervision, it is essential to consider metrics that reflect both point cloud density and accuracy. Inspired by \cite{cheng2022novel}, we define the density metric Radar Point Cloud Density ($RPCD$) as the proportion of supervision LiDAR points around which there exist radar points within a radius of $\delta_d=0.3m$, and the accuracy metric Radar Point Cloud Accuracy ($RPCA$) by the proportion of radar points of which there are LiDAR points within a radius of $\delta_a=0.5m$.

% Therefore, we follow \cite{cheng2022novel} to define the density metric Radar Point Cloud Density ($RPCD$) by the proportion of supervision LiDAR points of which there are radar points within a radius of $\delta_d=0.3m$, and define the accuracy metric Radar Point Cloud Accuracy ($RPCA$) by the number of clutter radar points of which there are no LiDAR points within a radius of $\delta_a=0.5m$.

% Since the final thresholding that transform 3D occupancy grids to radar point clouds may have great impact on the output point clouds, and brings a trade-off between the density and accuracy, we calculate and draw the curve of $RPCD$-$RPCA$ by setting different thresholds. When the threshold is higher, there will be less points in the radar point cloud so $RPCD$ may decrease, but the accuracy of each point may improve, so $RPCA$ will increase. On the contrary, when the threshold is lower, $RPCD$ may increase and $RPCA$ may decrease.

% 基于metrics的一些结果，和传统方法对比
\subsection{Experimental Results}
In the experiment, we employ our DenserRadar algorithm along with our ground truth generation pipeline to the dataset. Additionally, we adapt GAN-based radar point cloud detection algorithm RPD-Net associated with its ground truth generation pipeline\cite{cheng2022novel}. Moreover, we also implement two classical and widely-used CFAR-type algorithms CA-CFAR and OS-CFAR based on the resources provided within the K-Radar dataset.

The final thresholding process that converts 3D occupancy grids into radar point clouds may significantly influence the characteristics of the resulting point clouds. Therefore, the thresholds of each algorithm are adjusted to ensure that the average number of points in the generated point clouds remains approximately constant. Based on this standardization, the $RPCD$ and $RPCA$ metrics for each algorithm are calculated and presented in Table. \ref{Tab:compare}. 

It is clear that our algorithm demonstrates superior performance compared to RPD-Net, CA-CFAR and OS-CFAR, achieving higher values in both density $(RPCD)$ and accuracy $(RPCA)$ when the number of points are approximately the same. Since the RPD-Net is originally designed for processing Range-Doppler(RD) maps from radars with relatively low resolution, it may encounters challenges when applied to high-resolution 4D tensors in the K-Radar dataset. Similarly, the traditional CA-CFAR and OS-CFAR algorithms also have difficulty in maintaining both density and accuracy.

\begin{table}[!htbp]
    \centering
    \caption{Comparison of metrics for different algorithms when generating approximately a fixed number of points ($RPCD$/$RPCA$).}
    \resizebox{\linewidth}{!}{
    \begin{tabular}{ccccc}
        \toprule
        Algorithms &$\sim$5000 &$\sim$10000 &$\sim$20000 &$\sim$50000\\
        \midrule
        Our Method &\textbf{0.11}/\textbf{0.42} &\textbf{0.14}/\textbf{0.36} &\textbf{0.19}/\textbf{0.31} &\textbf{0.32}/\textbf{0.26} \\ 
        RPD-Net &0.02/0.14 &0.04/0.11 &0.08/0.08 &0.13/0.05 \\
        CA-CFAR &0.03/0.22 &0.05/0.18 &0.11/0.15 &0.15/0.08 \\
        OS-CFAR &0.04/0.16 &0.08/0.13 &0.15/0.07 &0.18/0.03 \\
        \bottomrule
    \end{tabular}}
    \label{Tab:compare}
\end{table}

\begin{figure*}
    \centering
    \includegraphics[width = \linewidth]{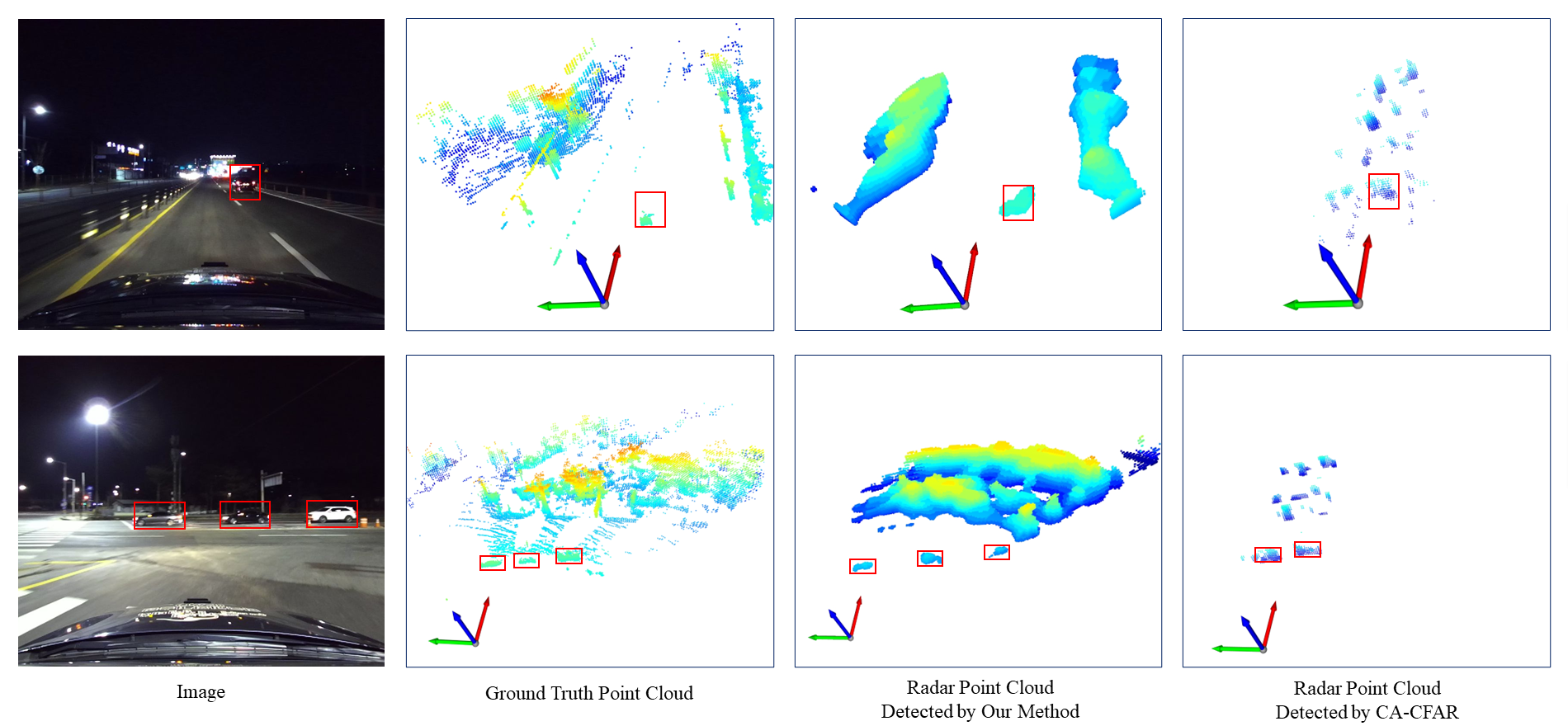}
    \caption{The qualitative point cloud comparisons between our DenserRadar algorithm and the CA-CFAR algorithm, accompanied by images and dense 3D occupancy ground truth point clouds for reference. Each arrow in the figures represents a length of 10 meters.}
    \label{Fig:results}
\end{figure*}

% TODO: 一张各算法效果对比图，2-3张自己效果和真值（LiDAR, image）对比图和分析
% TODO：考虑列个表，给出不同固定RPCD下，各算法的RPCA？
Furthermore, the qualitative point cloud comparisons between our DenserRadar algorithm and the CA-CFAR algorithm under two scenarios are shown in Fig. \ref{Fig:results}. The point clouds of dense 3D occupancy ground truth and the corresponding scenario images are also attached. The results illustrate that our algorithm shows good effect in detecting dynamic objects, such as surrounding vehicles, suggesting that the Doppler velocity information and intensity values are proficiently encoded by the DenserRadar network. However, static objects like surrounding vegetation may occasionally be missed or falsely detected due to their sparse and noisy radar reflections. Besides, as the LiDAR is mounted on the top of the vehicle while the 4D mmWave radar is at the front, the FOV of the radar may be limited, preventing the detection of objects located beyond the road boundaries present in the ground truth. In contrast, the point clouds detected by CA-CFAR—using the algorithm provided in the dataset that converts the 4D mmWave radar tensor from spherical to Cartesian coordinates—exhibit a constrained FOV, further limiting the radar's detection capabilities.

% 去掉一些核心环节后的结果
\subsection{Ablation Study}
To further validate the effectiveness of the specially designed modules in our algorithm, we perform a series of ablation studies to compare the performances of these four algorithms:

\begin{itemize}
    \item Our initially proposed DenserRadar algorithm (A).
    \item DenserRadar without cross-attention (B).
    \item DenserRadar with vanilla focal loss only at the end of the network (C).
    \item DenserRadar with ground truth generated by single frame LiDAR point cloud (D).
\end{itemize}

The outcomes of the ablation studies are demonstrated in Table. \ref{Tab:ablation}. The results clearly indicate that each of the modules under test contributes positively to the performance in nearly all the cases. Notably, the most impactful module is the integration of stitched multiple-frame LiDAR ground truth, which substantially enhances both the density and accuracy of the point clouds. The implementation of the weighted hybrid loss function also yields prominent improvement. While the cross-attention module generally performs well, there are some exceptions where input data noise may misleads the attention mechanism, leading to suboptimal performance.

\begin{table}[!htbp]
    \centering
    \caption{Ablation Study for different modules of our algorithm ($RPCD$/$RPCA$).}
    \resizebox{\linewidth}{!}{
    \begin{tabular}{ccccc}
        \toprule
        Algorithms &$\sim$5000 &$\sim$10000 &$\sim$20000 &$\sim$50000\\
        \midrule
        A &\textbf{0.11}/\textbf{0.42} &\textbf{0.14}/0.36 &\textbf{0.19}/\textbf{0.31} &\textbf{0.32}/\textbf{0.26} \\ 
        B &0.09/0.41 &0.12/\textbf{0.38} &0.17/\textbf{0.31} &0.26/0.25 \\
        C &0.06/0.34 &0.10/0.30 &0.15/0.29 &0.24/0.23 \\
        D &0.04/0.22 &0.06/0.17 &0.09/0.14 &0.13/0.09 \\
        \bottomrule
    \end{tabular}}
    \label{Tab:ablation}
\end{table}

\section{Conclusion}  \label{Sec:Conclusion}
In this paper, we introduce DenserRadar, a novel 4D mmWave radar point cloud detection network, along with an innovative pipeline for generating dense ground truth. The experimental results and ablation studies prove the effectiveness of our network architecture and ground truth generation methodology. This research has the potential to promote the perception and localization capabilities of autonomous driving systems, particularly in challenging corner-case scenarios.

In the future, we will further improve the radar point cloud detector by integrating generative models, such as diffusion models, and validate the enhanced detection performance through downstream perception and localization tasks.

%后续工作：用扩散模型尝试、进行下游实验

\bibliographystyle{IEEEtran} %声明选择的格式

\bibliography{ref}

\addtolength{\textheight}{-12cm}

\end{document}